\documentclass{article} 
\usepackage[final]{colm2026_conference}

\usepackage{microtype}
\usepackage{hyperref}
\usepackage{url}
\usepackage{booktabs}
\usepackage[most]{tcolorbox}
\usepackage{makecell}
\usepackage{wrapfig}
\usepackage{caption}
\usepackage{amsmath}
\usepackage{subcaption}
\usepackage{bbding}
\usepackage{graphicx}
\usepackage{multirow}
\usepackage{makecell}
\usepackage{enumitem}

\usepackage{lineno}

\definecolor{darkblue}{rgb}{0, 0, 0.5}
\hypersetup{colorlinks=true, citecolor=darkblue, linkcolor=darkblue, urlcolor=darkblue}

\title{Valley3: Scaling Omni Foundation Models for E-commerce}

\author{%
Zeyu Chen\textsuperscript{1,*} \quad
Guanghao Zhou\textsuperscript{2,*} \quad
Min Yang\textsuperscript{3,*} \quad
Qixiang Yin\textsuperscript{4} \quad
Ziwang Zhao\textsuperscript{3} \quad
\\[2.5pt]
\textbf{Huanjin Yao}\textsuperscript{3} \quad
\textbf{Pengjiu Xia}\textsuperscript{3} \quad
\textbf{Cen Chen}\textsuperscript{2} \quad
\textbf{Minghui Qiu}\textsuperscript{3,\ensuremath{\dagger}}
\\[4pt]
\textsuperscript{1}Tsinghua University \quad
\textsuperscript{2}East China Normal University
\\[2.5pt]
\textsuperscript{3}ByteDance \quad
\textsuperscript{4}Beijing University of Posts and Telecommunications
}

\begin{document}

\ifcolmsubmission
\linenumbers
\fi

\maketitle

\begingroup
\renewcommand{\thefootnote}{*}
\footnotetext{Equal contribution.}
\endgroup

\begingroup
\renewcommand{\thefootnote}{\ensuremath{\dagger}}
\footnotetext{Corresponding author: \texttt{minghuiqiu@gmail.com}}
\endgroup

\begin{abstract}
In this work, we present Valley3, an omni multimodal large language model (MLLM) developed for diverse global e-commerce tasks, with unified understanding and reasoning capabilities across text, images, video, and audio. A key feature of Valley3 is its native multilingual audio capability for e-commerce, developed by extending vision-language models to better support crucial audio-visual tasks, particularly in short-video scenarios. To achieve this, we carefully design a four-stage omni e-commerce continued pre-training pipeline, through which Valley3 progressively acquires audio understanding, cross-modal instruction-following, e-commerce domain knowledge, and long-context reasoning capabilities, ultimately evolving into an omni model for diverse e-commerce scenarios. Then, we further improve Valley3 through post-training to encourage long-chain reasoning with controllable reasoning modes, enabling one non-thinking mode and two distinct levels of thinking, thereby balancing inference efficiency in simple scenarios with deep reasoning for complex applications. Moreover, we equip Valley3 with agentic search capabilities to proactively invoke search tools and acquire task-relevant information for e-commerce deep research tasks. To comprehensively assess the capabilities of Valley3, we construct an omni e-commerce benchmark spanning 6 tasks. Experimental results show that Valley3 consistently outperforms strong baselines on our in-house and open-source e-commerce benchmarks, while remaining competitive on general-domain benchmarks. Our code and model weights are available at \url{https://github.com/bytedance/Valley}.
\end{abstract}

\section{Introduction} 

The e-commerce ecosystem is evolving at an extraordinary pace, with its complexity, scale, and dynamism challenging existing technological paradigms.
Across verticals such as customer service, search, recommendation, governance, content moderation, and after-sales support, systems must process massive volumes of multimodal data—including text, images, video, live streaming, and audio—in real time. Conventional deep learning models, limited by siloed architectures and uni-modal processing, struggle to satisfy the growing need for cross-modal understanding, long-tail demand fulfillment, and dynamic decision-making, especially amid ambiguous user intents, diverse product representations, and cross-lingual compliance requirements in global operations. The emergence of MLLMs~\citep{multimodal_llm_survey} creates a transformative opportunity to reshape this technological foundation.

However, while general-purpose models excel in open-domain tasks, we find them inadequate for e-commerce pipelines for three main reasons.
(1) \textbf{Limited perception and domain grounding}: General MLLMs lack fine-grained e-commerce knowledge, such as product taxonomies and platform regulations, and multimodal perception, particularly audio-visual understanding. Consequently, they are prone to errors and hallucinations in e-commerce tasks, especially in short-video and live-stream scenarios.
(2) \textbf{Reasoning–efficiency trade-off}: Many e-commerce tasks require complex multi-step reasoning, but current reasoning MLLMs often produce excessively long reasoning traces, leading to substantial inference overhead and making deployment difficult in latency-sensitive online systems.
(3) \textbf{Rapidly evolving e-commerce environments}: Rapid changes in products, policies, and user behaviors quickly outpace the parametric knowledge of MLLMs, which is largely fixed after training, limiting their ability to remain accurate and up to date in real-world settings.

These limitations motivate us to present \textbf{Valley3}, an efficient omni e-commerce foundation model for unified e-commerce understanding and reasoning. Building on the Valley series~\citep{luo2023valley,wu2025valley2}, Valley3 advances along three key aspects:

\textbf{Omni E-Commerce Capabilities}.
To address limited perception and domain grounding, Valley3 extends a pre-trained vision-language model by introducing audio as a primary modality, enabling unified text-audio-visual understanding. Through a four-stage e-commerce continued pre-training curriculum, it progressively develops from basic audio understanding to long-context cross-modal interactive reasoning, thereby strengthening multimodal perception and domain adaptation across diverse e-commerce scenarios.

\textbf{Controllable Reasoning Effort}. 
To balance reasoning capability and efficiency, we equip Valley3 with controllable reasoning effort, consisting of one non-thinking mode and two thinking modes. This design allows us to allocate reasoning depth according to task complexity, avoiding excessive computation on simple tasks while enabling deeper multi-step reasoning when needed. In this way, Valley3 improves deployment efficiency without sacrificing performance on complex e-commerce scenarios.

\textbf{Agentic Search Capabilities.}
To overcome the limitations of static knowledge, Valley3 is equipped with an agentic search mechanism through agentic post-training. It can autonomously formulate queries and invoke external search tools to retrieve real-time, task-relevant information, from regulatory updates to emerging market trends. This empowers the model to conduct deep, evidence-grounded research for knowledge-intensive, long-tail e-commerce challenges.

To benchmark Valley3, we also introduce the \textbf{EComm} benchmark, a comprehensive suite constructed from real-world e-commerce data pipelines. It provides robust evaluation across key dimensions, including Product Understanding, After-sales Experience, Search \& Recommendation, Livestream Content Analysis, Short Video Understanding, and Moderation \& Governance.
Extensive experiments demonstrate that Valley3 achieves excellent results on both our in-house and open-source e-commerce benchmarks while remaining competitive on general-purpose multimodal evaluations.

\section{Valley3 Architecture}
\subsection{Overview}
\begin{figure*}[!htbp]
  \centering
  \includegraphics[width=\linewidth]{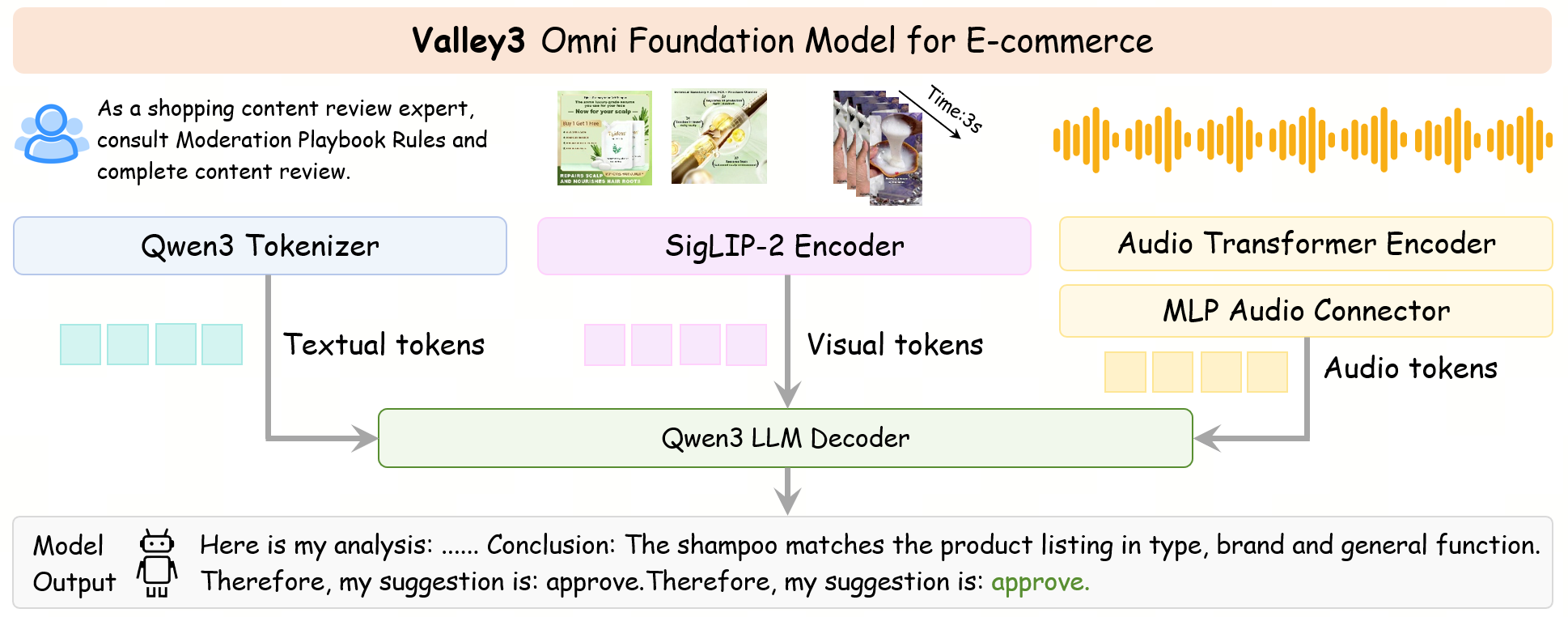}
  \caption{The architectural overview of Valley3. It is built upon the Qwen3-VL backbone and extends it with audio transformer for audio encoding. The audio embeddings are aligned to the visual-language backbone via an MLP-based connector, then concatenated with visual and text tokens into a unified input space, enabling omni-modal understanding.}
  \label{image:valley-arch}
\end{figure*}

Valley3 is a unified multimodal model that extends the Qwen3-VL backbone with native audio understanding capability derived from Qwen3-Omni, as shown in Figure~\ref{image:valley-arch}. The goal is to preserve the strong vision–language reasoning ability of Qwen3-VL (8B or 32B variants) while incorporating high-quality speech representations through a modular audio extension. The architecture follows a simple principle: visual tokens, text tokens, and projected audio tokens are mapped into a shared embedding space and jointly processed by a single decoder-only transformer.

\subsection{Vision-Language Backbone}
We adopt the Qwen3-VL family as the vision-language backbone. Qwen3-VL is a large-scale multimodal model that unifies image and text modeling within a decoder-only transformer architecture. It supports high-resolution visual inputs, multi-image interleaving, and strong instruction-following capability. In Valley3, we retain the original backbone of Qwen3-VL (vision encoder, multimodal projector, and decoder-only transformer) and its vision-language alignment mechanism. We extend the input token space to incorporate audio tokens that are aligned to the same hidden dimension.

\subsection{Audio Extension}
To enable audio understanding, Valley3 integrates the audio transformer (AuT) encoder from Qwen3-Omni into the Qwen3-VL backbone through a dedicated audio connector. 

Given a raw waveform, the AuT encoder outputs acoustic features $H^{audio} \in \mathbb{R}^{T \times I}$ ($I=2048$ for Qwen3-Omni). Since this dimensionality differs from the hidden size $H$ of the Qwen3-VL backbone ($H=4096$ for 8B and $H=5120$ for 32B VL backbone), we project the audio features into the shared embedding space using a three-layer MLP with an input dimension of $I$, hidden dimensions of 8192, and output dimensions of $H$, using GELU activations between layers and a final LayerNorm for stabilization.

The projected audio embeddings $Z^{audio} \in \mathbb{R}^{T' \times H}$ are concatenated with visual and textual embeddings to form a unified input sequence, which is processed by the unchanged decoder-only transformer. This embedding-level fusion strategy allows Valley3 to incorporate speech representations from AuT while preserving the original vision-language architecture.

To properly inject spatial and temporal context into this unified sequence, we adopt Time-aligned Multimodal Rotary Position Embedding (TM-RoPE) and follow their alignment strategy for processing multimodal audiovisual streams.

\section{Continual Pre-training}
\subsection{Continual Pre-training Recipe}

\begin{table}[h]
\centering
\caption{Training setup across different stages for Valley3.}
\label{tab:qwen3vl_training}
\scalebox{0.8}{
\setlength{\tabcolsep}{5pt}
\begin{tabular}{l l l c c}
\toprule
\textbf{Stage} & \textbf{Objective} & \textbf{Training} & \textbf{Token Budget} & \textbf{Sequence Length} \\
\midrule
S0 & Audio-Language Alignment & Audio Connector & $\sim$10B & 8,192 \\
S1 & Audio Instruction Fine-tuning & All & $\sim$ 1.5B & 8,192 \\
S2 & In-domain Continual Pre-Training & All & $\sim$1B & 32,768 \\
S3 & Rejection Sampling Fine-Tuning & All & $\sim$0.1B & 65,536 \\
\bottomrule
\end{tabular}}
\end{table}

\paragraph{Stage 0: Audio-Language Alignment.}
In Stage 0 (S0), we train the audio connector to align the audio representations with the language embedding space. The model is trained using large-scale multilingual audio-text pairs. The vision-language backbone is initialized with Qwen3-VL and kept frozen during this stage. The audio encoder is adopted from the AuT encoder of Qwen3-Omni. Only the audio connector is trained to project the acoustic features into the shared embedding space of vision-language model. This stage enables the model to learn a stable mapping from audio representations to the visual and textual embedding space while preserving the original multimodal capabilities.
    
\paragraph{Stage 1: Audio Instruction Fine-tuning.}
In Stage 1 (S1), we perform audio instruction fine-tuning to enhance audio understanding and reasoning abilities. Unlike S0, we jointly optimize all model components in S1. We curate a diverse audio instruction dataset spanning tasks such as audio question answering and sound event detection, encouraging the use of acoustic cues beyond transcription. To preserve vision-language capabilities and stabilize training, we mix in a small amount of general vision-language instruction data, which regularizes the model and mitigates drift from its original multimodal knowledge. Overall, S0 and S1 enable the model to acquire strong audio understanding while maintaining robust performance on existing vision-language tasks.

\paragraph{Stage 2: In-domain Continual Pre-Training.}
In Stage 2 (S2), we integrate multimodal e-commerce knowledge and restore general vision-language capabilities while preserving S1 audio proficiency. We train on $\sim$1B tokens with a 32k sequence length and a 2:1 ratio of in-domain to open-domain data. To ensure cross-modal alignment and balance domain-specific versus general performance, we fine-tune the proportions of image, text, and audio data. Furthermore, experience replay with downsampled audio data is employed to mitigate catastrophic forgetting, ultimately enabling unified cross-modal instruction following.

\paragraph{Stage 3: Rejection Sampling Fine-Tuning.}
In Stage 3 (S3), we enhance instruction-following and generalization by applying a model-guided rejection sampling strategy to the S2 dataset. Specifically, the S2 model filters the original data by discarding trivial fully correct samples and noisy highly incorrect samples to focus on those near the decision boundary. Concurrently, the context length is expanded to 64k. This strategy increases the density of informative signals without shifting the overall data distribution, thereby improving performance in both multimodal e-commerce and general domains.
\subsection{Continual Pre-training Data}
\begin{figure*}[!htbp]
  \centering
  \includegraphics[width=\linewidth]{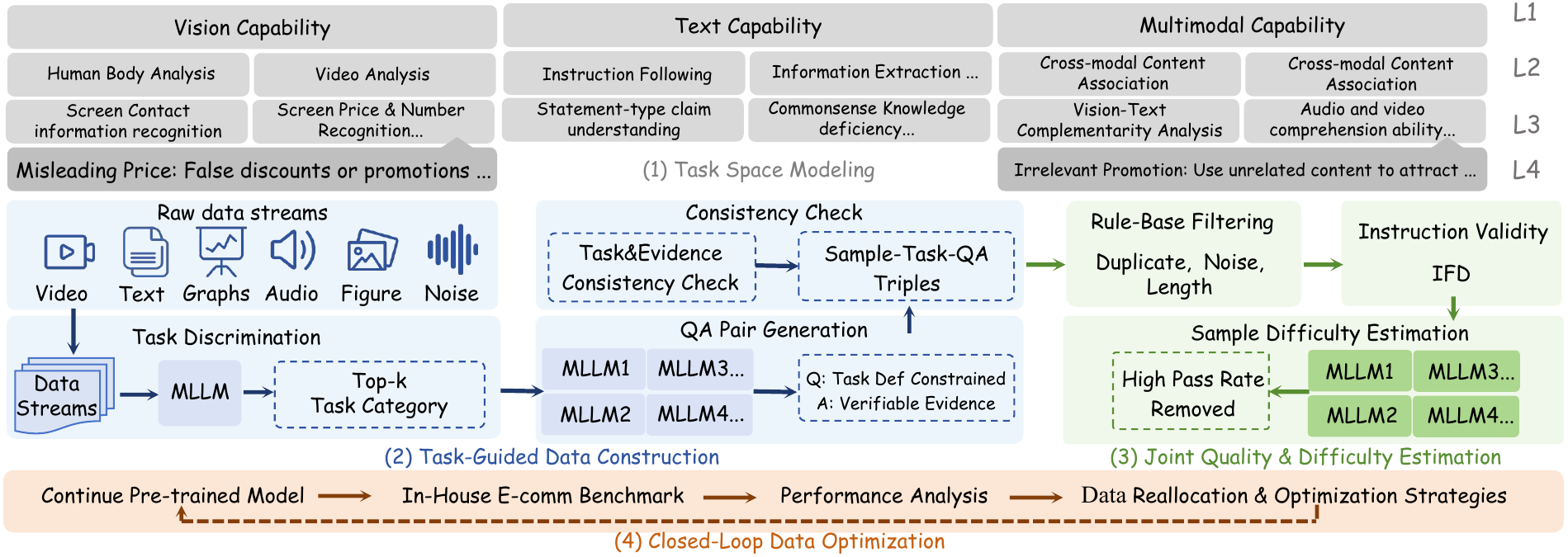}
  \caption{Overview of the proposed data construction framework for the e-commerce domain. The pipeline aligns the data distribution with the model’s capability boundary through four integrated modules: (1) Task Space Modeling; (2) Task-Guided Data Construction; (3) Joint Quality \& Difficulty Estimation; and (4) Closed-Loop Data Optimization.
  }
  \label{image:e-comm-cpt-data}
\end{figure*}

To meet the capability requirements of MLLMs in the e-commerce domain, we redesign the construction of training data from a distributional perspective to improve coverage and alignment with target capabilities.
Our framework explicitly models the task space and incorporates difficulty-aware data selection, enabling the data distribution to adaptively match the model’s evolving capability boundary. As illustrated in Figure~\ref{image:e-comm-cpt-data}, the overall pipeline integrates multiple components into a unified closed-loop optimization process. While our data construction is centered on e-commerce capabilities, we additionally retain a curated set of general-domain multimodal data to maintain broad generalization and reasoning abilities, with details deferred to Appendix~\ref{appendix:subsection-general-data}.

\paragraph{Task Space Modeling}
Current MLLM training typically mixes heterogeneous data sources, reducing capability learning to a data scaling problem and leading to unclear capability boundaries and inefficient data utilization. To address this, we model e-commerce capabilities as a set of atomic tasks $\mathcal{T}$ and explicitly annotate each sample with its task type. Under this formulation, capabilities are hierarchically refined from $L1$ to $L4$. $L1$ partitions capabilities by modality, such as vision, language, and audio. $L2$ groups capabilities with similar objectives and data distributions, such as image recognition and video analysis. $L3$ decomposes capabilities into minimally executable atomic units, such as logo recognition and frame localization, enabling flexible composition of complex tasks. $L4$ maps atomic capabilities to business rules to support core e-commerce tasks, such as product understanding and content moderation. This design offers two advantages. It organizes data along capability dimensions to make the training distribution explicit, and it provides structured constraints for data generation and filtering.
Detailed task definitions are given in Appendix~\ref{sec:appendix_ecom_tasks}, with the full capability taxonomy in Appendix ~\ref{appendix:subsection-e-commerce-data}.

\paragraph{Task-Guided Data Construction}
E-commerce multimodal data are highly heterogeneous and often contain sparse semantics, inconsistent fields, and domain-specific abbreviations, which makes fine-grained cross-modal alignment difficult.  To address this, we use the atomic task space $\mathcal{T}$ as an intermediate scaffold: an MLLM first maps each input to a top-$k$ set of relevant tasks, and a model pool then generates task-constrained question--answer pairs grounded in verifiable evidence. We further enforce both task consistency and evidence consistency, retaining only aligned task--question--answer triples and thereby reducing hallucination while improving supervision relevance.

\paragraph{Joint Quality \& Difficulty Estimation}
The data generated in the previous stage exhibits substantial variability in quality and learning value, ranging from low-level issues like redundancy and atypical length to deeper imbalances in instruction strength and reasoning reliability. To address these limitations, we propose a hierarchical filtering mechanism that jointly evaluates content usability, instruction effectiveness, and reasoning reliability. After discarding unusable samples via length filtering and duplicate removal, we introduce Instruction Following Difficulty (IFD) to measure the dependence of the output:
$\mathrm{IFD}(y\mid x)=\mathrm{PPL}(y\mid x)/\mathrm{PPL}(y)$,
Samples with low IFD, indicating limited training value, are removed. Finally, we estimate reasoning stability via cross-model accuracy across multiple MLLMs, pruning overly simple instances with high pass rates. This annotation-free process effectively yields training data with significantly higher information density.

\paragraph{Closed-Loop Data Optimization}
As model capabilities improve, easy examples dominate training signals, leaving difficult cases that define the performance ceiling underrepresented. We propose an iterative data refinement mechanism that targets these bottlenecks by performing fine-grained evaluation of atomic tasks on an e-commerce benchmark. By stratifying samples based on rollout outcomes, the mechanism prioritizes high-value examples while suppressing low-information ones. This continuous feedback loop dynamically recalibrates the data distribution, focusing supervision on underdeveloped capabilities to accelerate convergence on challenging tasks.

\section{Post-training}
\subsection{Post-training Training Recipe}
\begin{figure*}[!htbp]
  \centering
  \includegraphics[width=\linewidth]{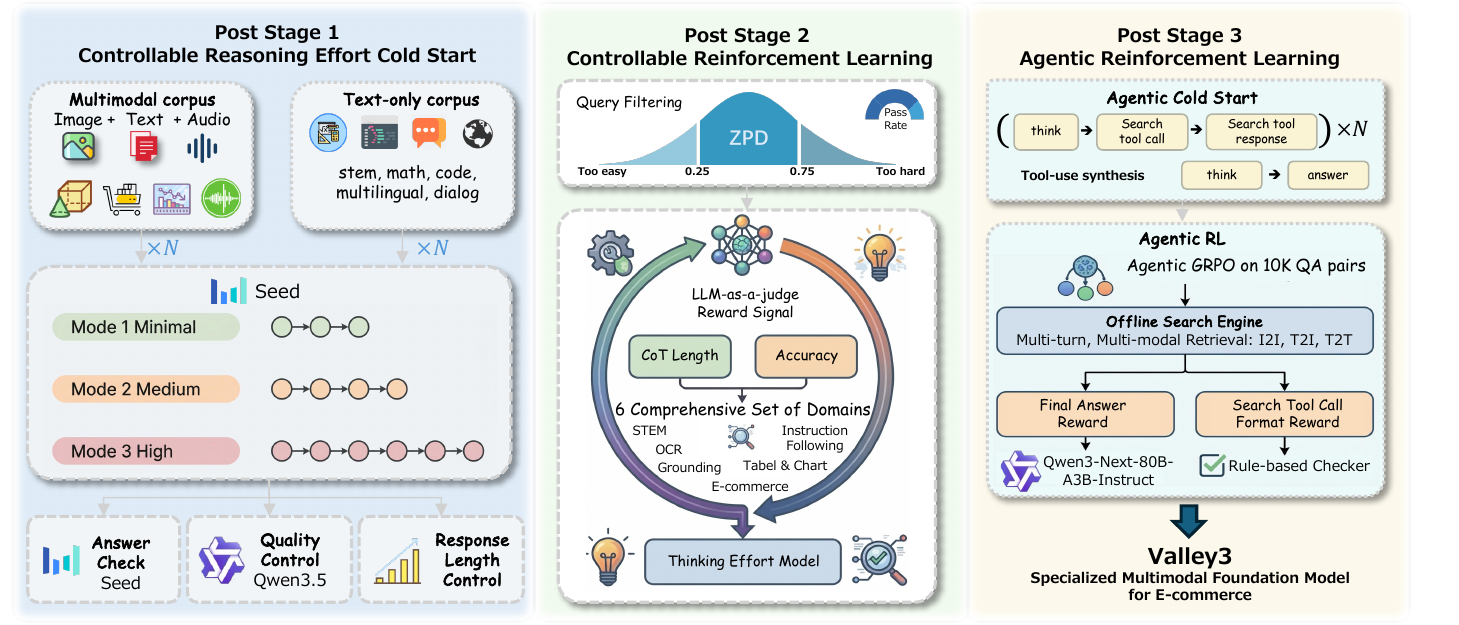}
  \caption{Overview of the proposed post-training recipe for Valley3. The pipeline enhances both e-commerce domain expertise and general reasoning capabilities through three progressive stages: (1) Cold Start for Controllable Reasoning; (2) Reinforcement Learning for Controllable Reasoning; and (3) Agentic Reinforcement Learning.}
  \label{image:valley-posttrain}
\end{figure*}

In the post-training stage, we equip Valley3 with two key capabilities: controllable reasoning effort and agentic search tool use. Controllable reasoning effort allows Valley3 to adapt to diverse e-commerce scenarios with varying latency constraints and reasoning-depth requirements. Agentic search further enables the model to handle information-intensive and time-sensitive e-commerce deep research tasks more effectively. As illustrated in Figure~\ref{image:valley-posttrain}, we adopt a three-stage fine-tuning recipe to achieve this. The details are described below.

\textbf{Post Stage 1: Cold Start for Controllable Reasoning.}
In the first stage, we enable Valley3 to adapt its reasoning process to a specified reasoning-effort level, covering both a no-thinking mode and multiple thinking modes with different reasoning depths. To this end, we synthesize and curate chain-of-thought data of varying lengths across domains and modalities, and conduct supervised fine-tuning so that the model learns the reasoning-length distribution associated with each reasoning-effort level.

\textbf{Post Stage 2: Reinforcement Learning for Controllable Reasoning.}
In the second stage, we further enhance model performance under different reasoning-effort levels via reinforcement learning. Specifically, we perform large-scale RL training over a diverse set of tasks, leveraging a unified LLM-as-a-Judge framework to provide reward signals. We further develop task-specific scoring rubrics that account for the unique characteristics of each task, thereby improving the accuracy and robustness of reward estimation.

\textbf{Post Stage 3: Agentic Reinforcement Learning.}
In the final stage, we conduct agentic RL on a carefully curated dataset to enhance Valley3’s capabilities in multi-turn agentic search. Through this stage, the model learns to proactively invoke external search tools, formulate intermediate queries, and iteratively refine its reasoning based on retrieved information.

\subsection{Post Stage 1: Controllable Reasoning Effort Cold Start}

At this stage, our primary objective is to equip the model with the ability to reason at different effort levels across both e-commerce and general-domain tasks. We consider three reasoning modes: a non-thinking mode with minimal effort, and two thinking modes with increasing reasoning depth, corresponding to medium and high effort levels. To realize this capability, we construct a cold-start dataset through a three-step pipeline: question-answer pair selection, chain-of-thought synthesis, and data filtering. This process produces high-quality supervision data aligned with different reasoning-effort levels, providing the foundation for subsequent training.

\textbf{QA pair selection.} We first collect data from both e-commerce and genera sources. For each question-answer pair, we prompt Valley3 Instruct to generate multiple long-chain reasoning solutions and assess their correctness.
We then filter out examples with high correctness rates, as they are typically already well mastered or relatively easy.

\textbf{CoT synthesis.} Given the selected QA pairs, we use Seed~\citep{seed2026seed1_8} to generate multiple candidate chain-of-thought trajectories spanning different reasoning depths. Specifically, we design three prompts corresponding to three reasoning modes, ranging from no thinking to medium and high reasoning effort.
Notably, the length of all generated CoT data is constrained within a predefined range, as controlling reasoning length is important for deploying thinking models under practical e-commerce latency constraints.
Then, we use Valley3 to rewrite the generated CoT trajectories in order to better align the data distribution with the model’s native generation patterns and mitigate catastrophic forgetting.

\textbf{Data Filtering.} To ensure high-quality reasoning data with the target reasoning lengths, we apply a multi-stage filtering pipeline.
(1) \textit{Final Answer Evaluation.} We use an LLM to verify whether the final answer matches the ground-truth answer.
(2) \textit{Reasoning Quality Control.} We assess the reasoning process and filter out trajectories with repetitive patterns or logical inconsistencies.
(3) \textit{Progressive Response Length.} For each question, we ensure that the chain-of-thought length increases progressively across the three reasoning-effort levels.

This stringent data production and curation process yields 1.1M CoT samples spanning three reasoning-effort levels, which are used to bootstrap advanced reasoning through SFT.

\subsection{Post Stage 2: Reinforcement Learning for Controllable Reasoning}

\textbf{Data Preparation.} For RL, we adopt a fine-grained, targeted difficulty selection strategy. Specifically, we use the cold-start model to generate multiple rollouts for each question under each of the three reasoning-effort levels. Motivated by the Zone of Proximal Development~\citep{zpd}, we retain examples whose rollout accuracy falls within the $0.25$--$0.75$ range, as they are expected to provide the most informative learning signals during RL post-training. This process yields distinct training subsets for different reasoning-effort levels, enabling targeted optimization under each setting. We balance the data volume across effort levels, resulting in 100K RL queries spanning a diverse range of tasks.

\textbf{Reward System.} We uniformly employ an LLM-as-a-Judge framework equipped with task-specific scoring rubrics to provide the accuracy reward $R_{\text{acc}}$. This mechanism is specifically formulated to reward both the appropriate scaling of CoT length and the accuracy of the reasoning process across different specified thinking-effort levels. We design different scoring rubrics for different task categories, which can be grouped into general-domain verifiable and non-verifiable tasks, and e-commerce tasks. For each task category, we use task-specific prompts to guide reward model in producing accurate and reliable reward signals. To ensure that the model follows the required response format, we further introduce a rule-based format reward $R_{\text{fmt}}$, which checks whether the generated response correctly contains the \texttt{\textless thinking\textgreater\textless/thinking\textgreater} tags. The final Stage-2 reward is a weighted combination of the two components: $R_{\text{stage2}} = 0.9R_{\text{acc}} + 0.1R_{\text{fmt}}$.

\textbf{RL Algorithm.} We employ Group Relative Policy Optimization (GRPO)~\citep{guo2025deepseek}, a method that consistently delivers robust performance gains while demonstrating remarkable generalizability across diverse model scales and architectures.

\subsection{Post Stage3: Agentic Reinforcement Learning}
Inspired by prior work on multimodal deep research~\citep{MM-DeepResearch,tongyidr}, we further enhance the agentic search capability of Valley3 through a two sub-stage training paradigm, enabling it to proactively invoke diverse search tools and integrate returned information into its reasoning process, thereby improving its ability to handle knowledge-intensive and dynamically evolving e-commerce tasks.

In the agentic cold-start stage, we generate and filter 15K search tool-use trajectories. Each trajectory follows a multi-turn interleaved format, i.e., \((\texttt{think} \rightarrow \texttt{search tool call} \rightarrow \texttt{search tool response}) \times N \rightarrow (\texttt{think} \rightarrow \texttt{answer})\). We then use these synthesized trajectories to perform supervised fine-tuning (SFT) on Valley3.

We then perform agentic multi-turn RL in a fully offline search environment. To enable efficient and scalable training, we build an offline retrieval engine over a large pre-indexed text--image corpus, supporting image-to-image, text-to-image, and text-to-text retrieval. The engine employs E5~\citep{wang2022text} for text retrieval and Jina-CLIP~\citep{koukounas2024jina} for multimodal retrieval. Since all training questions are answerable from this corpus, the framework eliminates the cost and latency of online search APIs while maintaining a realistic search experience.

During RL, we optimize the model with GRPO on 10K QA pairs, using two reward signals to improve answer quality and enforce the desired reasoning format: (1) \textbf{final answer reward} $R_{\text{acc}}$, evaluated by Qwen3-Next-80B-A3B-Instruct against the gold answer; and (2) \textbf{search tool-call format reward} $R_{\text{fmt}}$, computed by a rule-based checker to verify whether the response follows the required interleaved format. Similar to post stage 2, the overall reward is defined as: $R_{\text{stage3}} = 0.9R_{\text{acc}} + 0.1R_{\text{fmt}}$.

\section{Experiments}
\subsection{E-commerce Benchmark}
\subsubsection{EComm Benchmark Construction}
To evaluate model performance across diverse e-commerce scenarios, we assess Valley3 on both a newly constructed in-house benchmark and established open-source datasets.

\textbf{In-house Benchmark.} We build a general multimodal and multilingual e-commerce benchmark covering the full lifecycle of online commerce across platform-facing, merchant-facing, and user-facing settings. It comprises 6 fundamental tasks that capture the major functional scenarios in modern e-commerce: Product Understanding, After-sales Experience, Search \& Recommendation, Livestream Content Analysis, Short Video Understanding, and Moderation \& Governance. Detailed descriptions of these tasks are provided in Appendix \ref{sec:appendix_ecom_tasks}. 
Built on anonymized, real-world shop data, the benchmark primarily measures zero-shot accuracy. To reflect real business environments, our prompts integrate complex operational rules and require step-by-step reasoning. All prompts and labels were rigorously verified by internal experts to ensure high quality and avoid data contamination. Detailed descriptions of the EComm benchmark can be found in Appendix \ref{sec:appendix_ecomm_benchmark}.

\textbf{Open-source Benchmarks.} To ensure a comprehensive and fair evaluation, we also assess Valley3 on open-source benchmarks. We select three representative benchmarks with distinct focuses: ECInstruct~\citep{peng2024ecellm} for general e-commerce instruction following, ChineseEcomQA~\citep{chen2025chineseecomqascalableecommerceconcept} for fundamental e-commerce concepts and domain-specific QA, and MMECInstruct~\citep{ling2024captions} for diverse capabilities across seven tasks, such as product classification, substitute identification, and recommendation.

\subsubsection{EComm Benchmark Evaluation}

\begin{table*}[!htbp]
  \centering
  \caption{\textbf{Main Results.} To evaluate the effectiveness of Valley3, we compare it with baselines on both e-commerce and general tasks across text, image, video, and audio modalities.}
  \scalebox{0.65}{
  \setlength{\tabcolsep}{3pt}
  \begin{tabular}{c|c|ccc|ccc|c}
    \toprule
    Capabilities & Benchmark & \makecell{Valley3 32B\\Instruct} & \makecell{Valley3 32B\\Think} & \makecell{Qwen3-VL\\32B} & \makecell{Valley3 8B\\Instruct} & \makecell{Valley3 8B\\Think} & \makecell{Qwen3-VL\\8B} & \makecell{Qwen3-Omni\\30B-A3B} \\
    \midrule
    \multirow{6}{*}{\makecell{E-commerce\\(Ours)}} & Product Understanding & \textbf{77.2} & 76.8 & 69.7 & 76.7 & 76.1 & 64.6 & 63.7 \\
    & After-sales Experience & 78.4 & \textbf{80.7} & 77.3 & 79.5 & 79.1 & 76.7 & 71.6 \\
    & Search \& Recommendation & 76.9 & \textbf{77.2} & 76.3 & 74.6 & 75.9 & 70.9 & 68.2 \\
    & Livestream Content Analysis & \textbf{84.5} & 84.1 & 76.2 & 75.8 & 75.5 & 65.1 & 69.6 \\
    & Short Video Understanding & 83.0 & \textbf{84.5} & 82.7 & 81.9 & 82.1 & 80.2 & 83.3 \\
    & Moderation \& Governance & \textbf{82.7} & 82.5 & 73.6 & 76.7 & 76.3 & 66.8 & 64.9 \\
    \midrule
    \multirow{3}{*}{\makecell{E-commerce \\(Open)}} & ECInstruct & 48.4 & \textbf{49.4} & 47.0 & 45.4 & 46.2 & 45.4 & 47.3 \\
     & ChineseEcomQA & \textbf{66.2} & 64.3 & 65.8 & 61.2 & 60.3 & 61.9 & 64.5 \\
    & MMECInstruct & 62.6 & \textbf{63.0} & 60.8 & 59.3 & 60.4 & 59.9 & 59.4 \\
    \midrule
    \multirow{3}{*}{Image} & MMMU & \textbf{74.2} & 73.4 & 72.8 & 69.3 & 66.1 & 68.6 & 68.7 \\
    & MMMU-Pro$_{10c}$ & 49.0 & \textbf{62.9} & 47.4 & 44.6 & 52.4 & 40.2 & 45.2 \\
    & HallusionBench & 60.2 & \textbf{61.3} & 59.8 & 55.9 & 59.6 & 58.8 & 59.7  \\
    \midrule
    \multirow{2}{*}{Language} & SuperGPQA & 44.4 & \textbf{49.6} & 43.5 & 33.4 & 36.1 & 32.1 & 44.6  \\
    & PolyMath & 35.3 & \textbf{50.0} & 34.9 & 30.9 & 39.5 & 30.7 & 32.0  \\
    \midrule
    \multirow{2}{*}{Video} & VideoMMMU & \textbf{68.7} & 67.9 & 68.3 & 61.2 & 60.9 & 61.7 & 63.6 \\
     & MLVU & \textbf{59.1} & 57.6 & 56.3 & 55.6 & 54.7 & 54.1 & 58.9 \\
    \midrule
    \multirow{3}{*}{Audio}
     & MMAU-Sound & \textbf{81.7} & 79.6 & - & 78.1 & 75.7 & - & 78.4 \\
     & MMAU-Music & \textbf{75.8} & 74.3 & - & 72.8 & 72.5 & - & 75.2 \\
     & MMAU-Speech & 75.7 & \textbf{83.8} & - & 74.8 & 82.9 & - & 80.5 \\
     \midrule
     \multirow{2}{*}{Search}
     & EcomBench & - & \textbf{44.0} & 32.0 & - & 36.0 & 25.0 & - \\
     & MMSearch & - & \textbf{67.2} & 44.4 & - & 68.7 & 37.4 & - \\
    \bottomrule
  \end{tabular}}
  \label{tab:Main_Results_thinking_instruct}
  \vskip -0.1in
\end{table*}

Table \ref{tab:Main_Results_thinking_instruct} presents the comparative evaluation of our proposed Valley3 models against the Qwen3-VL and Qwen3-Omni series on both the in-house and open-source e-commerce benchmarks, while additional comparisons with other baseline models are provided in Appendix~\ref{sec:appendix_baseline_comparison}. Overall, Valley3 demonstrates substantial improvements in domain-specific tasks while maintaining strong generalization capabilities across diverse e-commerce scenarios. Specifically, Valley3 makes remarkable progress in Product Understanding, Livestream Content Analysis, and Moderation \& Governance tasks, achieving an average absolute improvement of over 7\% compared to the baselines. On open-source e-commerce benchmarks, Valley3 32B achieves consistent improvements and Valley3 8B maintains comparable performance, demonstrating robust generalization without overfitting to in-house data.

\subsection{General Benchmark Evaluation}
To comprehensively evaluate the general and omni-modal capabilities of Valley3, we conduct experiments on a diverse set of open-source benchmarks spanning multiple modalities, including text, image, video, and audio. We use VLMEvalKit~\citep{duan2024vlmevalkit} and EvalScope~\citep{evalscope_2024} as the evaluation frameworks for these open-source benchmarks. The results are shown in Table~\ref{tab:Main_Results_thinking_instruct}.

\textbf{Text understanding.} We evaluate Valley3’s language understanding and reasoning abilities on the SuperGPQA~\citep{pteam2025supergpqascalingllmevaluation} and PolyMATH~\citep{wang2025polymath} benchmarks, focusing on cross-domain generalization and multilingual reasoning. At 32B, Valley3 Instruct already surpasses Qwen3-VL 32B on both SuperGPQA and PolyMath, with scores of 44.4 and 35.3 versus 43.5 and 34.9. Valley3 Think further widens the gap, reaching 49.6 on SuperGPQA and 50.0 on PolyMath, substantially ahead of all baselines. The same trend holds at 8B, clearly exceeding both Valley3 Instruct and Qwen3-VL 8B. These results show that Valley3 has stronger language reasoning ability overall, and that the Thinking variant brings especially large gains on reasoning-intensive benchmarks.

\textbf{Image Understanding.}
We assess the model’s image understanding capabilities on MMMU~\citep{DBLP:conf/cvpr/YueNZ0LZSJRSWYY24}, MMMU-Pro$_{10c}$~\citep{DBLP:conf/acl/YueZNW0T0Y000CN25}, and HallusionBench~\citep{DBLP:conf/cvpr/GuanLWXLL0CHYM024} benchmarks, focusing on multi-disciplinary understanding and reasoning, and hallucination. The results indicate that Valley3 Instruct maintains and even enhances general visual reasoning alongside its domain-specific tuning. At 32B, Valley3 consistently outperforms Qwen3-VL on image benchmarks. Valley3 Instruct leads on MMMU with 74.2, while Valley3 Think achieves the best results on MMMU-Pro$_{10c}$ and HallusionBench with 62.9 and 61.3. A similar pattern is observed at 8B. Valley3 remains stronger overall, and reasoning brings larger gains on benchmarks that require more complex visual reasoning.

\textbf{Video Understanding.} We evaluate temporal understanding capability of Valley3 on VideoMMMU~\citep{DBLP:journals/corr/abs-2501-13826} and MLVU~\citep{DBLP:conf/cvpr/0001S0WLXQYXZH025}, which cover multi-disciplinary expert-level videos and long-video understanding, respectively. We uniformly sample 8 frames per video as the default setting. Benefiting from large-scale e-commerce video data during training, Valley3 exhibits strong video understanding capability. Notably, Valley3 32B-Ins achieves top-tier performance, scoring 68.7 on VideoMMMU and 59.1 on MLVU.

\textbf{Audio Understanding.} We evaluate the model’s audio understanding and reasoning performance across sound, speech, and music using the MMAU benchmark~\citep{DBLP:conf/iclr/SakshiTKSSNDGM25}. Valley3 demonstrates strong omni-modal capability in this domain. Valley3 32B-Ins achieves 81.7 on MMAU-Sound and 75.8 on MMAU-Music, outperforming Qwen3-Omni 30B-A3B (78.4 and 75.2, respectively). In addition, Valley3 32B-Think attains 90.7 on MMAU-Speech, substantially exceeding Qwen3-Omni 30B-A3B (80.5). These results highlight Valley3’s strong audio capabilities across diverse audio domains.

\textbf{Agentic Search.} We evaluate Valley3’s agentic search capability on EcomBench~\citep{min2025ecombench} and MMSearch~\citep{jiang2024mmsearch}, which asses retrieval and evidence integration ability. Valley3 8B-Think achieves 36.0 on EcomBench and 68.7 on MMSearch, surpassing Qwen3-VL 8B by 11.0 and 31.3 points, respectively. These results demonstrate the effectiveness of our agentic training pipeline for search and evidence integration.

\section{Conclusion}
In this paper, we present Valley3, a series of omni large language models for e-commerce domain. Through large-scale e-commerce CPT, Valley3 extends vision-language models with native audio capabilities, enabling unified understanding and reasoning across text, images, video, and audio for complex e-commerce tasks. Valley3 also introduces controllable reasoning to balance reasoning performance and inference latency across different scenarios, while incorporating agentic search to actively retrieve task-relevant information in rapidly evolving e-commerce environments. Experimental results show that Valley3 performs strongly on both e-commerce and general benchmarks, demonstrating the potential of domain-oriented omni models for real-world e-commerce applications.



\bibliography{colm2026_conference}
\bibliographystyle{colm2026_conference}

\clearpage

\appendix

\section{Related Work}

\subsection{Omni Large Language Models}
Multimodal Large Language Models (MLLMs) have rapidly evolved from text-image understanding to unified processing, representation and reasoning across text, vision, audio and diverse data types~\citep{minicpmv4_5,qwen3omni,li2025baichuan_omni_1_5}. One primary line of research focuses on modality alignment and unified architectures~\citep{fu2025vita1_5,liu2025ola}. These works typically project heterogeneous inputs into a shared language model backbone via continuous encoding or modality-specific adaptors, enabling cross-modal generation and understanding within a single framework~\citep{luo2025openomni,geng2025longvale,li2025megrez}. Another prominent direction explores temporal synchronization and dynamic perception. These models establish robust connections across visual, audio, and subtitle tracks to handle streaming data and long-context video events~\citep{chen2023vast,tan2025omni_video,qwen35omniblog}.
However, despite strong performance on general tasks, state-of-the-art MLLMs still suffer from limitations in specialized domains requiring deep domain expertise, particularly in e-commerce. To address this, we propose Valley3, an omni-modal MLLM tailored for e-commerce that supports diverse real-world e-commerce applications.

\subsection{E-commerce Foundation Model}

E-commerce foundation models have been widely applied across various e-commerce scenarios~\citep{herold2025ellama,yang2025llm_driven_ecommmerce}. Recent efforts have primarily focused on adapting these models to the e-commerce domain, enhancing their foundational representation abilities and complex reasoning skills~\citep{qiu2026Thinking_Broad_Acting_Fast,wang2026shoppingbench} for applications such as zero-shot product classification~\citep{ling2025captions,maria2025compass_v3}, multimodal recommendation~\citep{zhang2026onemall}, and conversational agents~\citep{li2024ecomgpt}. More recently, emerging research has explored agentic retrieval mechanisms. In contrast to static retrieval, methods such as \citep{kim2026agenticshop, li2026kuaisearch, liu2025taosearchemb} employ autonomous agentic search to interact with dynamic commodity environments, supporting real-time updates of inventory and item availability. However, current e-commerce foundation models~\citep{luo2023valley,wu2025valley2,dong2025taosr1} are mostly limited to text-vision modalities and struggle to handle rapidly growing complex formats like short videos and livestreams, which require joint reasoning across audio, visual, and textual streams. Our proposed Valley3 addresses this gap as an omni-modal MLLM tailored for e-commerce, integrating multi-modal understanding, agentic search, and controllable reasoning to solve real-world e-commerce tasks.

\section{Details of Fundamental E-commerce Tasks}
\label{sec:appendix_ecom_tasks}
As mentioned in the main text, our in-house e-commerce benchmark defines 6 fundamental multimodal tasks that capture the major functional scenarios in modern e-commerce. The detailed definitions and challenges of these categories are as follows:

\textbf{Product Understanding.} Given multimodal product information such as images, titles, descriptions, and attributes, the model needs to accurately understand the product itself. Representative challenges include recognizing diverse logos, identifying fine-grained product attributes, classifying products into appropriate categories, verifying image-text consistency, and distinguishing visually similar products in the e-commerce domain.
    
\textbf{After-sales Experience.} The model is required to assist both users and merchants throughout the full post-purchase after-sales service process. Typical tasks include answering after-sales related inquiries, handling returns, exchanges and warranty claims, conducting standardized after-sales compliance education for merchants, clarifying liability determination for transaction and product quality disputes and formulating and matching compliant and reasonable compensation plans.

\textbf{Search \& Recommendation.} This task evaluates the model's ability to identify high-quality content and accurately assess query-product relevance. The core challenge lies in aligning user intent with product semantics, especially when queries are ambiguous, underspecified, or expressed across multiple modalities.

\textbf{Livestream Content Analysis.} The model is required to understand multimodal content in live-stream commerce, including event localization, static frame detection, and product consistency verification. The key challenge lies in jointly modeling temporal dynamics, conversational context, and product-related signals in highly dynamic shopping environments.

\textbf{Short Video Understanding.} The model is required to understand short e-commerce videos that present products through demonstrations, reviews, or promotional narratives. Representative tasks include video--cart product consistency, cart brand consistency checking, and video comment classification. The main challenge lies in capturing temporal visual cues, aligning them with textual or spoken information, and inferring product-related semantics in dynamic, real-world scenarios.
    
\textbf{Moderation \& Governance.} The model is required to identify unsafe, misleading, or policy-violating content in e-commerce scenarios. This includes product pornography recognition, financial currency recognition, pirated broadcast detection, livestream sexual innuendo recognition, and the detection of other risky or non-compliant behaviors. The main challenge lies in recognizing subtle violations that often require joint reasoning over both visual and textual evidence.

\section{Details of the EComm Benchmark}
\label{sec:appendix_ecomm_benchmark}
\subsection{Construction of the EComm Benchmark}
Our in-house e-commerce benchmark is constructed from real-world commercial data collected in production environments, rather than being adapted from existing open-domain academic benchmarks. The construction pipeline consists of the following stages.

\textbf{Data Collection and Sampling.} Candidate samples are collected from diverse e-commerce scenarios, covering the six major categories of e-commerce tasks described in Section \ref{sec:appendix_ecom_tasks}. Stratified sampling is applied to ensure balanced coverage across task categories, modalities, and difficulty levels.

\textbf{Question-Answer Generation.} Evaluation questions are constructed based on the collected commercial data and practical e-commerce requirements through a combination of expert-designed templates and MLLM-assisted generation. All reference answers are manually annotated and verified by domain experts to ensure correctness, consistency, and evaluation reliability.

\textbf{Quality Control.} Each sample undergoes multiple quality assurance procedures, including automatic filtering, expert review, and consistency checking. Ambiguous, duplicated, or low-quality samples are removed to improve the reliability of the benchmark.

\subsection{Characteristics of the EComm Benchmark} 
Compared with existing open-domain multimodal benchmarks, our proposed EComm presents several unique challenges arising from real-world e-commerce scenarios, including the following representative examples.

\textbf{Fine-grained product understanding.}
Many tasks require distinguishing subtle differences among visually similar products, such as material, specifications, style, or packaging variants. Correct predictions rely on fine-grained attribute recognition rather than coarse object recognition.

\textbf{Complex policy reasoning and violation attribution.}
E-commerce governance policies vary across product categories and business scenarios, while policy violations are often implicit (e.g., misleading claims or evasive promotions). Models must reason over multimodal evidence to identify violations while maintaining a low false-positive rate.

\textbf{Robust understanding of noisy livestream content.}
Livestream scenarios are characterized by noisy visual and audio signals, multiple simultaneously presented products, ambiguous references, and imperfect visual quality. Models are required to perform long-horizon multimodal grounding and temporal reasoning while remaining robust to irrelevant information.

These characteristics make the benchmark substantially more representative of real-world e-commerce applications and considerably more challenging than existing open-domain benchmarks. Table \ref{tab:ecomm_comparison} highlights the differences between our benchmark and existing e-commerce benchmarks.

\begin{table}[h]
\centering
\caption{Comparison of existing e-commerce benchmarks and our EComm.}
\label{tab:ecomm_comparison}
\small
\renewcommand{\arraystretch}{1.15}
\begin{tabular}{p{2.2cm}p{4.6cm}p{6.0cm}}
\toprule
\textbf{Dimension} & \textbf{Existing Benchmarks} & \textbf{EComm (Ours)} \\
\midrule
\textbf{Data Source} & Web-scraped or synthetic data with isolated scenarios & Real-world shop data collected from production environments \\
\textbf{Task Coverage} & Focused on individual tasks or narrow application settings & Covers the complete e-commerce lifecycle through six fundamental task categories \\
\textbf{Modality} & Primarily text and image & Omni-modal, including text, image, video, and audio \\
\textbf{Reasoning Complexity} & Object recognition, classification, or basic QA & Fine-grained product understanding, policy reasoning, and long-horizon multimodal grounding \\
\bottomrule
\end{tabular}
\end{table}

\section{Evaluation Prompts}
For reproducibility and fair evaluation, we directly utilize the default prompts from standard evaluation frameworks where applicable. For benchmarks requiring specific configurations, the customized prompts are detailed below.

\begin{tcolorbox}[
  title=ECInstruct,
  fonttitle=\bfseries,
  colframe=black,
  colback=white,
  toptitle=1mm,
  bottomtitle=1mm,
  top=2mm,
  verbatim,
]
Analyze the user's review text and determine the overall sentiment expressed, then choose the corresponding sentiment option from the provided list (A: very positive, B: positive, C: neutral, D: negative, E: very negative) based on the identified sentiment.

- Output Format

Your output must strictly adhere to the following JSON format.

\{

    "thinking\_process": "xxx(A brief thought process)",
    
    "answer": "xxx(only the final answer, without any extra text, such as your chosen option or yes/no)"
    
\}

<query>

<options>

\end{tcolorbox}

\section{Prompts in Post-training Stages}
\label{sec:appendix_rl_prompts}
\subsection{Training data verification}
For annotation correctness verification, the verification prompt is not a single fixed prompt. Instead, we employ a set of task-specific verification prompts, which are adjusted according to the nature and evaluation requirements of different tasks. A template is provided below.

\begin{tcolorbox}[
  title=Prompt for annotation correctness verification,
  fonttitle=\bfseries,
  colframe=black,
  colback=white,
  toptitle=1mm,
  bottomtitle=1mm,
  top=2mm,
  verbatim,
]
You are a strict evaluator.

Task: Determine whether the model's output answer is CONSISTENT with the provided ground-truth answer.

You will be given: the question (with images), the model output, and the ground-truth answer.

Rules:

1) Focus on the final answer/claim in the model output, not the style.

2) If the model output clearly agrees with gt\_answer in meaning => check='yes'.

3) If it contradicts, answers a different thing, or is unclear/insufficient to confirm => check='no'.

4) If gt\_answer is a short label (e.g., Yes/No/A/B), treat it as the required final label.

You MUST call the save\_check function.
\end{tcolorbox}

\subsection{Inference of Valley3-Think model}
\begin{tcolorbox}[
  title=System prompt for controllable reasoning effort,
  fonttitle=\bfseries,
  colframe=black,
  colback=white,
  toptitle=1mm,
  bottomtitle=1mm,
  top=2mm,
  verbatim,
]
You are a helpful assistant.

Reasoning effort: {mode}

Reasoning policy:

- Use the specified reasoning effort as an internal guide for how much analysis to do before answering, with the reasoning enclosed within <thinking> and </thinking>.

- The response generated after </thinking> MUST strictly follow the user's instructions and required output format.
Reasoning effort levels:

- minimal: Disable internal reasoning for this effort. Output an empty <thinking>\texttt{\textbackslash n}</thinking> before the response.

- medium: Perform internal reasoning, using clear step-by-step thinking and verifying important constraints before responding.

- high: Perform more thorough internal reasoning, consider multiple possible interpretations, alternatives, and edge cases, and carefully validate the final answer before responding.
\end{tcolorbox}
\clearpage

\subsection{GRPO Training: LLM-as-Judge prompt}
\begin{tcolorbox}[
  title=Prompt for the LLM-as-a-Judge module during GRPO,
  fonttitle=\bfseries,
  colframe=black,
  colback=white,
  toptitle=1mm,
  bottomtitle=1mm,
  top=2mm,
  verbatim,
]
You are an AI assistant tasked with evaluating whether a model response is correct, given a question and a ground truth answer.

Your judgment must follow these principles:

1. Evaluate the question, the ground truth answer, and the model response holistically before making a decision.

2. Your output must be strictly either "Yes" or "No".

3. Judge correctness based on factual consistency with the ground truth answer, not stylistic similarity.

4. If the model response is a more specific form of the ground truth answer, it is correct.

5. If the model response contains all essential information from the ground truth answer and adds only minor, factually correct details, it is correct.

6. If the model response contradicts, alters, or omits critical parts of the ground truth answer, it is incorrect.

7. For numerical answers, treat equivalent values in different units or equivalent mathematical forms as correct.

8. For names, check whether the core identity is correct; extra correct name details are acceptable.

9. If the question is a yes/no question, the model response must match the correct yes/no judgment to be correct.

10. If the model response is empty, irrelevant, refuses to answer, or does not directly answer the question, it is incorrect.

IMPORTANT: If the response is truncated due to excessive length, it shall be judged as incorrect.

Additional task-specific rules:

11. The ground truth answer may be a full solution process rather than a short answer.

12. If the problem is a standard question with a definite final answer (for example, a computation, multiple-choice, fill-in-the-blank, or short-answer question), only the final conclusion matters.
    
\quad - The reasoning steps do not need to match the ground truth.
    
\quad - If the final answer is correct, judge "Yes" even if the intermediate steps differ or are omitted.
    
\quad - If the final answer is wrong or missing, judge "No" even if some intermediate reasoning appears similar.
    
13. If the problem is a proof-based question, evaluate whether the key proof steps and core logical structure are consistent with the ground truth.
    
\quad - Exact wording, notation, ordering of minor details, or stylistic differences do not matter.
    
\quad - The model response may use a different but valid presentation, as long as the essential proof strategy and critical logical steps are present.
    
\quad - If the key steps are missing, logically invalid, or materially different from the ground truth proof, judge "No".
    
14. Do not penalize harmless differences in expression, formatting, or level of detail.

15. Focus only on correctness relative to the ground truth, not on elegance, completeness beyond the key points, or writing quality.

Output format: Yes or No
\end{tcolorbox}
\clearpage

\section{Additional Baseline Comparison}
\label{sec:appendix_baseline_comparison}
To provide a more comprehensive evaluation, we further include two proprietary general-purpose models (GPT-5 and Gemini-3 Pro) as additional baselines. Detailed comparison results are presented in Table~\ref{tab:additional_proprietary_baselines}. The Valley3 32B models remain highly competitive with these proprietary models across the e-commerce benchmarks.

In particular, Valley3 achieves competitive performance on ChineseEcomQA, demonstrating strong domain reasoning capability while preserving robust generalization to open-domain e-commerce tasks. Moreover, Valley3 achieves performance comparable to Gemini-3 Pro despite the latter being a substantially larger proprietary multimodal reasoning model. Overall, these results further validate the effectiveness of our training pipeline and demonstrate that Valley3 remains competitive with state-of-the-art proprietary models, particularly on domain-specific e-commerce tasks.

\begin{table*}[!htbp]
\centering
\caption{
Comparison with proprietary large-scale models on e-commerce benchmarks. The best result in each row is highlighted in \textbf{bold}.
}
\label{tab:additional_proprietary_baselines}
\resizebox{\textwidth}{!}{
\begin{tabular}{c|c|cccc}
\toprule
E-commerce
&
Benchmark
& \makecell{Valley3 32B\\Instruct} & \makecell{Valley3 32B\\Think} & GPT-5 & \makecell{Gemini-3\\Pro} \\
\midrule

\multirow{6}{*}{\makecell[l]{Ours}}
& Product Understanding & 77.2 & 76.8 & 77.9 & \textbf{80.1} \\

& After-sales Experience & 78.4 & 80.7 & \textbf{81.6} & 78.1 \\

& Search \& Recommendation & 76.9 & \textbf{77.2} & 74.3 & 74.0 \\

& Livestream Content Analysis & \textbf{84.5} & 84.1 & 78.9 & 83.3 \\

& Short Video Understanding & 83.0 & \textbf{84.5} & 82.9 & 80.3 \\

& Moderation \& Governance & \textbf{82.7} & 82.5 & 80.0 & 77.4 \\
\midrule
\makecell[c]{Open}
& ChineseEcomQA & 66.2 & 64.3 & 63.9 & \textbf{67.2} \\

\bottomrule
\end{tabular}
}
\end{table*}

\section{Ablation Study}
\subsection{Training Strategy}
\subsubsection{Continual Pre-training Stage}
A key challenge in training omni-modal models is aligning representations from different modalities. After integrating the audio module and training audio connector in S0, we investigate how the allocation of CPT training data across different stages influences the model's capabilities in each modality. We hypothesize that a staged training approach outperform a single-stage strategy that mixes all AQA (audio question answering) and VQA (visual question answering) data.

We compare two distinct training strategies following an initial audio alignment stage S0:
\begin{itemize}
    \item \textbf{Two-Stage Strategy (S1 AQA \& S2 VQA):} In S1, the model is trained on approximately 2.3 million audio-language samples to align the audio and language representations. During this stage, the visual modules are frozen. To prevent catastrophic forgetting of visual-language capabilities, a small amount of image-text data is included. The model from S1 is then trained on a dataset of approximately 2.4 million VQA samples, comprising both in-domain and open-source data. All model parameters are unfrozen. A small amount of AQA data is replayed to maintain audio capabilities.
    \item \textbf{Single-Stage Strategy (Mixing AQA and VQA data):} The model is trained from the S0 checkpoint on a mixed dataset containing all 2.3 million audio samples and 2.4 million VQA samples simultaneously. All module parameters are unfrozen throughout this single stage.
\end{itemize}

\textbf{Experiment results: }
The ablation experiments are conducted on Valley3 8B model. As shown in Table \ref{tab:ablation_1}, the two-stage training strategy outperforms the single-stage method in all tasks. It improves in-domain VQA by 13.55\%, open-domain VQA by 3.4\%, and audio tasks by 1.4\% on MMAU.
\begin{table}[h]
\centering
\caption{Comparison of two-stage vs. single-stage training strategies.}
\label{tab:ablation_1}
\scalebox{0.8}{
\setlength{\tabcolsep}{5pt}
\begin{tabular}{l c c c}
\toprule
\textbf{Stage} & \textbf{In-domain VQA} & \textbf{Open-domain VQA} & \textbf{MMAU AQA} \\
\midrule
S1 AQA \& S2 VQA & \textbf{67.16} & \textbf{46.72} & \textbf{76.8} \\
Mixing AQA and VQA data & 53.61 & 43.32 & 75.4 \\
\bottomrule
\end{tabular}}
\end{table}

\textbf{Conclusion: }Mixing AQA and VQA data in the early stages of training appears to create conflicting optimization goals, preventing the model from effectively learning both audio features and complex visual-language reasoning. By decoupling these tasks, the two-stage strategy first establishes a robust audio-language alignment. Subsequently, training on VQA data on this well-aligned foundation allows the model to acquire higher-order VQA capabilities more effectively, leading to superior overall performance.

\subsubsection{CPT Data Composition}
Prior work has demonstrated that incorporating text-only data during the finetuning of VLMs can enhance their performance on VQA tasks. However, the effect of this strategy on models with an additional audio modality has not been thoroughly investigated. We examine the impact of introducing text-only data during S2 on a omni-modal model, specifically evaluating the trade-offs between VQA and AQA capabilities.

We conduct an ablation study based on the checkpoint of S1 (8B). Throughout all experiments in this section, the audio module's parameters are kept frozen to isolate the effects of data composition on the LLM. We compare three distinct data composition strategies for S2 finetuning:
\begin{itemize}
    \item \textbf{VQA Only (Baseline):} The model is finetuned exclusively on in-domain VQA data. This serves as our performance baseline.
    \item \textbf{VQA + Text (1:1):} In addition to the VQA data, we introduce 100k samples of text-only data, mixed in a 1:1 ratio with the VQA samples.
    \item \textbf{VQA + Text + AQA (1:1:1):} Building upon the second setting, we further mix in 100k audio data samples, resulting in a 1:1:1 ratio across the three data types.
\end{itemize}

\textbf{Experiment results: } As shown in Table \ref{tab:ablation_2}, introducing text-only data creates a significant trade-off. While it substantially boosts VQA performance (in-domain VQA improving 9.1\%, open-domain VQA improving 2.4\%), it severely impairs audio capability (MMAU AQA dropping 8.7\%). Re-introducing AQA data leads to a partial recovery. Audio performance recovers close to baseline, but this comes with a slight degradation in VQA scores.
\begin{table}[h]
\centering
\caption{The impact of data composition in S2 on VQA and AQA performance.}
\label{tab:ablation_2}
\scalebox{0.8}{
\setlength{\tabcolsep}{5pt}
\begin{tabular}{l c c c}
\toprule
\textbf{Stage} & \textbf{In-domain VQA} & \textbf{Open-domain VQA} & \textbf{MMAU AQA} \\
\midrule
VQA Only & 67.16 & 46.72 & \textbf{76.8} \\
VQA + Text & \textbf{76.27} & 49.11 & 68.1 \\
VQA + Text + AQA & 73.58 & \textbf{49.29} & 76.2 \\
\bottomrule
\end{tabular}}
\end{table}

\textbf{Conclusion: }The data composition in S2 is critical for balancing multi-modal capabilities. While introducing pure text data is a highly effective strategy for improving VQA performance, it poses a substantial risk of degrading audio understanding when audio-specific supervision signals are absent. Reintroducing a balanced amount of audio data can mitigate this degradation and restore audio performance while retaining most of the VQA gains. This highlights the necessity of maintaining a principled data balance during multi-modal finetuning to prevent catastrophic forgetting and achieve a well-rounded model.

\subsection{Effectiveness of CPT Data Filtering}
To purify the noisy unsupervised instruction data, we compute sample perplexities using the Valley3 base model. By applying specific thresholds, with PPL ranging from 2 to 7 and Instruction Following Difficulty (IFD) ranging from 0.4 to 0.7, we filter for high-information-density instructions rather than random sampling. We evaluated this on an in-house Moderation \& Governance task:

As shown in Table \ref{tab:data_filter_comparison}, with a fixed training data volume of 60k samples, the model trained on IFD-filtered data achieves 66.21\% accuracy, compared with 60.70\% for the model trained on randomly sampled data, validating the effectiveness of IFD-based data selection.

\begin{table}[!htbp]
\centering
\caption{Comparison between filtered and randomly sampled training data.}
\label{tab:data_filter_comparison}
\small
\begin{tabular}{c|c}
\toprule
\textbf{Backbone} & \textbf{ACC} \\
\midrule
Training on 6w-filter & \textbf{0.6621} \\
Training on 6w-random & 0.6070 \\
\bottomrule
\end{tabular}
\end{table}

\subsection{Effectiveness of Agentic RL Training}
We further isolated the contribution of the agentic RL stage on search benchmarks.

As shown in Table \ref{tab:stage_comparison} below, while SFT provides marginal gains over the base model, the agentic RL stage acts as the primary catalyst for performance, yielding substantial absolute improvements of +10.0 on EcomBench and +16.4 on MMSearch post-SFT.

\begin{table}[!htbp]
\centering
\caption{Performance comparison on search benchmarks across different training stages.}
\label{tab:stage_comparison}
\small
\begin{tabular}{c|cc}
\toprule
\textbf{Training Stage} & \textbf{EcomBench} & \textbf{MMSearch} \\
\midrule
Base & 32.0 & 44.4 \\
+ SFT &	34.0 (+2.0)	& 50.8 (+6.4) \\
+ RL & \textbf{44.0 (+10.0 over SFT)} & \textbf{67.2 (+16.4 over SFT)}\\
\bottomrule
\end{tabular}
\end{table}

\subsection{Thinking Effort}

\begin{wrapfigure}{l}{0.50\textwidth} 
  \centering
  \includegraphics[width=\linewidth]{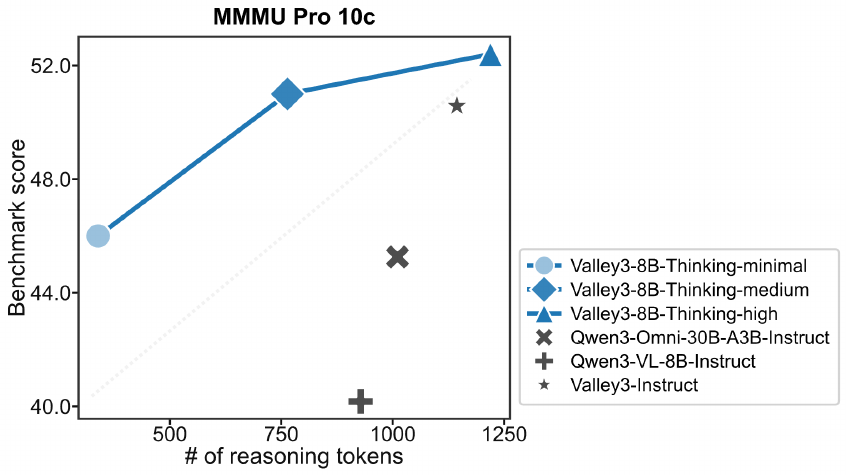}
  \caption{Performance of Valley3-8B-Thinking across varying reasoning effort levels on MMMU-Pro$_{10c}$.}
  \label{fig:thinking_effort_exp}
\end{wrapfigure}

While large language models exhibit strong reasoning capabilities, their performance is often constrained by fixed computational budgets, which hinders their adaptability to diverse business requirements in e-commerce scenarios. A core design principle of our Valley3-8B-Thinking model is its ability to scale reasoning effort on demand, enabling flexible tradeoffs between computational cost and task performance. To rigorously validate this capability, we conduct a systematic ablation study on the MMMU-Pro$_{10c}$ benchmark, evaluating model performance across a spectrum of reasoning effort configurations.

\textbf{Experiment results: } As shown in Figure \ref{fig:thinking_effort_exp}, Valley3-8B-Thinking consistently improves with increased reasoning effort, exhibiting a clear monotonic trend. Under the minimal setting, the model already achieves 46.0 accuracy with only 339.1 reasoning tokens, indicating strong efficiency. Scaling to medium effort further improves accuracy to 51.0, while the high-effort setting reaches the best performance of 52.4. Notably, the high-effort configuration also surpasses strong baselines, including Valley3-8B-Instruct , Qwen3-VL-8B-Instruct, and Qwen3-Omni-30B-A3B-Instruct, demonstrating the effectiveness of our effort-aware reasoning design.

\textbf{Conclusion: }Valley3-8B-Thinking exhibits improved performance with increased reasoning effort, offering efficient low-latency inference for e-commerce scenarios and state-of-the-art accuracy that outperforms larger models. The minimal effort setting strikes a compelling balance between accuracy and efficiency, making it highly suitable for latency-critical e-commerce applications. Meanwhile, the high effort configuration achieves state-of-the-art performance on MMMU-Pro$_{10c}$ and even outperforming larger baseline models. 

\section{Training Data}
\subsection{General Data}
\label{appendix:subsection-general-data}
To maintain the model's foundational generalization and reasoning abilities, we incorporate general data from open-source datasets such as FineVision~\citep{wiedmann2025finevisionopendataneed}, OpenMMReasoner~\citep{zhang2025openmmreasonerpushingfrontiersmultimodal}, and diverse audio sources. For vision-language data, we apply fine-grained filtering across global deduplication, visual dependency filtering, and data difficulty screening to create a high-quality subset. Similarly, for audio data, we collect paired audio-text samples across multiple languages and tasks (e.g., speech recognition, audio question answering, and emotion recognition) and enhanced quality through deduplication and audio-text consistency filtering. This comprehensive dataset ensures robust alignment, broad coverage of content, and improved general understanding across modalities.

\subsection{E-commerce Data}
\label{appendix:subsection-e-commerce-data}
Table~\ref{appendix:table-cpt-e-commerce-data} presents the categories at each level from $L1$ to $L3$ in the e-commerce atomic capability taxonomy used in CPT data construction, along with their corresponding relationships, to characterize the hierarchical structure and organization of different capabilities.

Based on this capability taxonomy, we further construct the CPT training corpus while carefully preventing data leakage from the evaluation benchmark. Specifically, we adopt three complementary strategies. (1) De-duplication. We de-duplicate text, image, and audio data to remove duplicate and near-duplicate content from the training corpus, reducing overlap with evaluation data. (2) Source-level partitioning. Evaluation product IDs, item IDs, and livestream IDs are excluded from the training pool before preprocessing. Since text, image, video, and audio samples associated with the same e-commerce entity share these identifiers, this filtering simultaneously removes related multimodal content, mitigating indirect contamination even when the exact evaluation samples are absent from the training set. (3) Task-format disentanglement. The instruction formats and question styles used for CPT differ from those in the benchmark, reducing overfitting to benchmark-specific prompts and evaluation patterns. As a result, the model achieves consistent improvements on our in-house e-commerce benchmarks while maintaining competitive performance on multiple open-domain benchmarks.

\begin{table}[!htbp]
\centering
\caption{Details of the e-commerce training data atomic capability taxonomy from $L1$ to $L3$.}
\resizebox{\textwidth}{!}{
\begin{tabular}{c|c|l}
\toprule
\textbf{L1 Capability Domain}          & \textbf{L2 Capability Category}                           & \textbf{L3 Atomic Capability}                       \\ \midrule
\multirow{33}{*}{Vision Capability}    & \multirow{10}{*}{Image Recognition}                       & Product Category Recognition                        \\
                                       &                                                           & Logo Recognition                                    \\
                                       &                                                           & Brand Recognition                                   \\
                                       &                                                           & Small Object Detection                              \\
                                       &                                                           & Similarity Comparison                               \\
                                       &                                                           & QR Code Recognition                                 \\
                                       &                                                           & Product detail recognition                          \\
                                       &                                                           & Recorded broadcast recognition                      \\
                                       &                                                           & Sticker Recognition                                 \\
                                       &                                                           & AIGC Image Recognition                              \\ \cmidrule(lr){2-3}
                                       & \multirow{4}{*}{Video Analysis}                           & Dynamic content duration estimation                 \\
                                       &                                                           & Superimposed video detection                        \\
                                       &                                                           & Static/Looping Frame Detection                      \\
                                       &                                                           & Scene detection                                     \\ \cmidrule(lr){2-3}
                                       & \multirow{5}{*}{OCR}        & Screen Contact information recognition              \\
                                       &                                                           & Screen Price \& Number Recognition                  \\
                                       &                                                           & Screen Text Recognition                             \\
                                       &                                                           & Screen Website/URL Recognition                      \\
                                       &                                                           & Product Packaging Text Recognition                  \\ \cmidrule(lr){2-3}
                                       & \multirow{3}{*}{Visual grounding/Counting}                & Object Localization                                 \\
                                       &                                                           & Temporal Grounding                                  \\
                                       &                                                           & Object Counting                                     \\  \cmidrule(lr){2-3}
                                       & \multirow{5}{*}{Human Body Analysis}                      & Face Localization and Attribute Recognition         \\
                                       &                                                           & Body Part Detection                                 \\
                                       &                                                           & Human Pose/Contour Recognition                      \\
                                       &                                                           & Before-After Effect Analysis                        \\
                                       &                                                           & Body Impurity/Blemish Recognition                   \\ \cmidrule(lr){2-3}
                                       & \multirow{6}{*}{Visual Quality Assessment}                & Image Clarity/Quality Assessment                    \\
                                       &                                                           & Abnormal/blurry image detection                     \\
                                       &                                                           & Duplicate screen detection                          \\
                                       &                                                           & Watermark and mosaic detection                      \\
                                       &                                                           & Lighting detection                                  \\
                                       &                                                           & Digital Tampering/Editing Trace Detection           \\  \midrule
\multirow{14}{*}{Text Capability}      & \multirow{6}{*}{Semantic Understanding}                   & Statement-type claim understanding                  \\
                                       &                                                           & Intent understanding                                \\
                                       &                                                           & NER, Named Entity Recognition                       \\
                                       &                                                           & Commonsense Knowledge deficiency                    \\
                                       &                                                           & Multilingual understanding                          \\
                                       &                                                           & Comparative Language Understanding                  \\ \cmidrule(lr){2-3}
                                       & \multirow{2}{*}{Instruction Following}                    & Output Format Compliance                            \\
                                       &                                                           & Output restrictions follow                          \\ \cmidrule(lr){2-3}
                                       & \multirow{3}{*}{Information Extraction}                   & Domain-specific Text Recognition                    \\
                                       &                                                           & Text Brand Recognition                              \\
                                       &                                                           & NER, Named Entity Recognition                       \\ \cmidrule(lr){2-3}
                                       & \multirow{3}{*}{\makecell{Harmful and Sensitive \\ Language Detection}} & Insulting/Abusive/Discriminatory Language Detection \\
                                       &                                                           & Negative Body Image Language Detection              \\
                                       &                                                           & Sexual Innuendo/Explicit Language Detection         \\ \midrule
\multirow{5}{*}{Multimodal Capability} & \multirow{2}{*}{Vision-Text Consistency}                  & Visual-Claim-to-Listing Consistency                 \\
                                       &                                                           & Visual-to-ASR Consistency                           \\ \cmidrule(lr){2-3}
                                       & \multirow{3}{*}{Cross-modal Content Association}          & Vision-Text Complementarity Analysis                \\
                                       &                                                           & Audio and video comprehension ability               \\
                                       &                                                           & Overall Marketing Judgment \\ \bottomrule                        
\end{tabular}}
\label{appendix:table-cpt-e-commerce-data}
\end{table}

\end{document}